\documentclass[journal]{IEEEtran}
\usepackage{graphicx} % more modern
\usepackage{subfigure} 
\usepackage{algorithm}
\usepackage{algorithmic}
\usepackage{amsmath,amssymb,amsgen,amsbsy,amsopn,bm,calc,epsfig,multicol,graphicx,color,psfrag}
\usepackage{wrapfig}

\def\x{{\mathbf{x}}}

\def\N{{N}}

\def\EP{\text{EP}}

\def\ZEP{{Z_{\EP}}}
\def\g{{\boldsymbol{\alpha}}}
\def\S{{\boldsymbol{\Lambda}}}

\def\backm{\backslash m}
\def\cav{q^{\backm}(\x)}
\def\til{p^{\backm}(\x)}
\def\tili{p^{\backm, \ell}(\x)}

\def\E{\mathbb{E}}
\def\vmu{\boldsymbol{\mu}}
\def\vSigma{\boldsymbol{\Sigma}}

\usepackage{hyperref}

\begin{document}
\title{Probabilistic Time of Arrival Localization}

\author{Fernando~P\'{e}rez-Cruz,
        Pablo~M.~Olmos,
        Michael~Minyi~Zhang,
        and~Howard~Huang% <-this % stops a space
\thanks{F. P\'{e}rez-Cruz is with the Swiss Data Science Center at ETH Z\"{u}rich (email: fernando.perezcruz@sdsc.ethz.ch). P. M. Olmos is with the Signal Processing Group at Universidad Carlos III de Madrid (email: pamartin@ing.uc3m.es). M. M. Zhang is with the Department of Computer Science at Princeton University (email: mz8@cs.princeton.edu). H. Huang is with Nokia Bell Labs (email: howard.huang@nokia-bell-labs.com)}%
%\thanks{}%
%\thanks{}%
%\thanks{}%
\thanks{This work has been funded in part by the Spanish Goverment under Grant TEC2016-78434-C3-3-R, in part by Comunidad de Madrid under Grants IND2017/TIC-7618, IND2018/TIC-9649, Y2018/TCS-4705, and in part by the European Research Council (ERC) through the European Union’s Horizon 2020 research and innovation program under Grant 714161.}
}

\markboth{}%
{}
\maketitle

\begin{abstract}
In this paper, we take a new approach for time of arrival geo-localization. We show that the main sources of error in metropolitan areas are due to environmental imperfections that bias our solutions, and that we can rely on a probabilistic model to learn and compensate for them. The resulting localization error is validated using measurements from a live LTE cellular network to be less than 10 meters, representing an order-of-magnitude improvement.
\end{abstract}


% \fi
\IEEEpeerreviewmaketitle

\section{Introduction}\label{Intro}

Geo-localization comes in many different flavors and requirements, so achieving centimeter-level accuracy might be straightforward in some cases, while achieving less than ten meters error seems unthinkable in other cases. For example, today we can achieve \textit{cm}-level accuracy indoor using ultra-wideband (UWB) systems \cite{Gunes16}. For outdoors we can rely on GPS with real time kinematic (RTK) positioning. But UWB is not scalable to an arbitrary number of users nor is RTK applicable to localization in metropolitan areas. %, so they can only provide solutions to niche localization applications. 
Meter-level accuracy can be achieved using fingerprinting of signal strength from Wi-Fi and cellular signals, as well as fingerprinting of the earth magnetic field, further assisted by inertial motion units \cite{Su16}, but these solutions seem unlikely to scale beyond specific indoor scenarios. Moreover, it is unclear how often the fingerprint database would require recalibration or how sensitive it is to device mismatch \cite{Lui11}. 
%
%The localization accuracy in smart phones today using Wi-Fi fingerprinting is measured in tens of meters and their usefulness requires a human to visually integrate its surroundings (examples of this are provided in the supplementary material). In short, today we have many non-scalable ad-hoc localization solutions that provide centimeter to meter level accuracy, while scalable universal localization errors are, at best, measured in the tens of meters. 

We are interested in a geo-localization system that is {\em scalable} (i.e., can serve any number of users) and {\em universal} (i.e., works indoors and outdoors). Having highly accurate, scalable and universal localization is crucial in practice, like in public safety scenarios where we may need accurate localization \cite{george2010distressnet}. 
% cite - public safety: “Distressnet: A wireless ad hoc and sensor network architecture for situation management in disaster response,” IEEE Commun. Mag., 2010.
%- IoT: “Efficient Multi-Sensor Localization for the Internet-of-Things,” IEEE Signal Process. Mag., 2018
%- crowd sensing: “Inhomogeneous Poisson sampling of finite-energy signals with uncertainties in Rd ,” IEEE Trans. Signal Process., 2016

%, hence we specifically focus on opportunistic localization with cellular networks designed for efficient data communications (e.g. 4G, LTE and 5G) that are available worldwide and that can serve the entire population of users for any potential application. 
To achieve this, we consider the localization problem using Time of Arrival (ToA) to understand the major limitations of ToA's accuracy and what changes we may introduce to improve it. ToA localization with cellular networks is hardly a novel idea \cite{Rappaport96}. The LTE standard has a specific Position Reference Signal (PRS) for downlink ToA localization. Most simulation studies show that ToA localization is quite accurate \cite{threeGPP_2015}.  These simulations achieve errors below 50 meters 90\% of the time and below 20 meters 50\% of the time, which should make ToA comparable, if not better, than Wi-Fi fingerprinting solutions in today's phones. If we add small cells, this error drops to single-digit meter level accuracy. 

%Currently, PRSs are only used by some networks providers in the USA to supplement GPS to meet the FCC localization mandate for 911 calls \cite{FCC_2015}. %The FCC requires that 40\% of the calls made to 911 from a cellular phone be localized within 50 meters (this will raise to 80\% by 2021). The accuracy of ToA in life LTE networks tends to be larger than 50 meters for most indoors calls (hundreds of meters in many cases). The FCC mandate is met due to GPS, which provides the requisite level of accuracy for most of the calls made outdoors. Or, in other words, if ToA were more accurate than Wi-Fi fingerprinting, we would use an application for smart phones that supplements GPS positioning instead.  

However, there is a significant gap between what simulations models could theoretically achieve and what we actually achieve. These differences are typically written off as unavoidable imperfections of the real LTE networks. Furthermore, it is common belief that if we move in the direction that the existing models propose--more bandwidth, more resolution, more hearability, etc.--localization accuracy will also improve in real cellular networks. However, we posit that by accounting for different sources of ToA error, we can in fact achieve accurate localization.

In this paper, we revisit ToA localization from a machine learning perspective, in which the environment needs to be learned and compensated for in order to provide accurate localization. This perspective allows us to improve localization with wireless networks to an extent that would be, at least, competitive with Wi-Fi fingerprinting solutions. Our starting point is a probabilistic model for ToA in which we can disambiguate each error source. 

To accomplish this, we propose a Bayesian generative model that allows us to understand the nature of each individual error and account for them in a principled manner. %We rely on expectation propagation (EP) \cite{Minka01a} for fast approximate inference. 
Our proposed work improves a previous idea for localization \cite{PerezCruz16} by proposing a computationally efficient algorithm \cite{Minka01a} and by accounting for miscalibration of the infrastructure nodes. We illustrate the benefits of our probabilistic algorithm using measurements from real localization networks, instead of relying on simulated data and propose changes to localization can be implemented in software over existing communication networks. 

%cite the following papers
%“Soft Range Information for Network Localization,” IEEE Trans. Signal Process., 2018
%“A mathematical model for wideband ranging,” IEEE J. Sel. Topics Signal Process., 2015
%“Ranging with ultrawide bandwidth signals in multipath environments,” Proc. IEEE, 2009
%“Network Navigation with Scheduling: Error Evolution,” IEEE Trans. Inf. Theory, 2017
%“Blind Selection of Representative Observations for Sensor Radar Networks,” IEEE Trans. Veh. Technol., 2015
%“Network Experimentation for Cooperative Localization,” IEEE J. Sel. Areas Commun., 2012
%“Experimental characterization of diversity navigation,” IEEE Syst. J., Mar. 2014

\section{Sources of Error}\label{sec2}
%ToA localization geometry is fairly straightforward. 
In ToA localization, we have a series of synchronized access points (APs) at known locations that simultaneously transmit their positioning signals. The devices within range measure the ToA of the positioning signals from all the APs within reach\footnote{Throughout the paper, we measure time and distance in meters using the speed of light to convert one into the other. We do not clutter our notation by dividing or multiplying by $c$. We use meters instead of nanoseconds, because we find them more intuitive.}:
\begin{equation}\label{ToA}
\text{ToA}_j-\tau=||\x-\bar{\x}_j||\qquad j=1,\ldots, N
\end{equation}
where $||\x||$ is the L2-norm of vector $\x$. These ToAs are measured with respect to the arrival time at the device of the positioning signal from the reference AP. Without loss of generality, we assume that the first AP is the reference, i.e. $\text{ToA}_1=0$. $\tau$ represents the unknown time of flight of the positioning signal from the reference AP to the device. $\x$ is the {\em unknown location} of the device and $\bar{\x}_j$ is the {\em known location} of AP $j$. This nonlinear system of equations can be solved to provide 2D (or 3D) localization if the device hears at least 3 (or 4) APs \cite[and the references within]{Fischer14,Guvenc09}. Most algorithms rely on the Time Difference of Arrival (TDoA) equations: $\text{TDoA}_j=||\x-\bar{\x}_j||-||\x-\bar{\x}_1||\ j=2,\ldots, N$, in which the reference AP equation is used to cancel out $\tau$. We rely on the ToA equations in \eqref{ToA}, instead of the (TDoA) equations, because estimating $\tau$ decorrelates the errors between the ToA estimates \cite{Sathyan10}. ToA localization performance depends critically on measurement errors. If the models in use are not tuned to deal with these errors, they can provide inaccurate localization \footnote{We note that there are current work in TDOA localization that can take into account outside sources of error \cite{vaghefi2015cooperative,ricciato2018position} though for the scope of this paper, we will compare with other TOA methods. Moreover, we may also incorporate complementary signals in conjunction with the ToA measurements \cite{coluccia2017hybrid,tomic2018robust,tomic2018target}.}. 
Below, we have rewritten the system of equations in \eqref{ToA} to account for each source of error: % that are present in the estimation of the ToA:
\begin{equation}\label{ToA_errors}
\text{ToA}_{ji}-\tau_i-w-n_{ji}-s_j-\delta_{T_j}=||\x_i-(\bar{\x}_j+\delta_{\bar{\x}_j})||+\gamma_{ji},
\end{equation}
where $ i $ indexes the devices and $ j $ indexes the APs, and 
%where we identify the device with the sub-index $i$ and sub-index $j$ for the AP, and
\begin{itemize}
	\item $w$ represents the quantization error. In current LTE, the ToAs are quantized to 32.56ns (9.77 meters). $w$ is uniformly distributed between $\pm5$ meters. %This value is expected to be at least half in future standards.
	\item  $n_{ji}$ is the unbiased estimation error %of the $\text{ToA}_{ji}$%
	 due to thermal noise. %If the communication is line of sight and a continuous matched filter is used for estimating the positioning reference, this error is Gaussian with a variance given by the Cram\'{e}r-Rao lower bound \cite{Dardari09}. %For example, using 10MHz signals and having 20dB of signal to noise ratio, the standard deviation of this error is about 1 meter.
	\item $s_j$ is the synchronization error of the APs. This error depends on the algorithm for synchronizing the APs. %For example, %for APs synchronized 
	 %the GPS standard says that 95\% of the cases with respect to UTC has error less than 40ns. %We could, for example, model it as an unbiased Gaussian with a standard deviation of 7 meters. In our data, we see synchronization errors significantly lower than the one reported by the GPS standard. 
	These errors are geographically clustered, so when measuring the ToA with respect to the reference AP, part of the synchronization error cancels out.
	\item $\gamma_{ji}$ is the non-line-of-sight bias.  %This value depends on the power delay profile between the device and AP, and the signal bandwidth, which is a statistical characterization of the additional travel time compared to the LOS distance  \cite{Medbo09}.% In cellular communications, this error can be as large as a few microseconds, which leads to biases measured in hundreds of meters.% and even kilometers.   
	$\delta_{T_j}$ represents the error in calibrating the delays in the AP.  %$\delta_{T_j}$ and $\delta_{\bar{\x}_j}$ represent biases due to AP miscalibration. %, because the time the signal takes from the baseband unit until it is transmitted by the radio-head is non-negligible. 
	$\delta_{\bar{\x}_j}$ represents the error from when the GPS unit does not record the AP's antenna position and from converting latitude, longitude and height to Cartesian coordinates, under the assumption that the Earth is a sphere. 
	%One, the GPS unit that collects the AP location might not record the antenna position. %(and height)which may lead to errors of a few meters. 
	%Two, when we convert latitude, longitude and height into Cartesian coordinates \cite{Vincenty75}, we assume the Earth is a sphere, leading to a few meters of error.% for typical cellular network distributions. 
\end{itemize}
\begin{figure}[h]
	\includegraphics[width=0.5\linewidth]{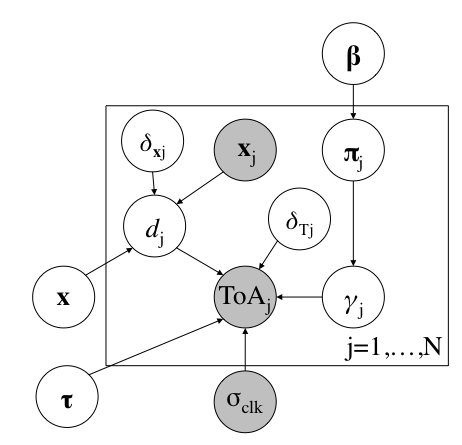}
	\centering
	\caption{Graphical Model for ToA geo-localization.}
	\label{fig1}
	%\vspace*{-10pt}
\end{figure}
%\begin{wrapfigure}{R}{0.4\textwidth}
%\end{wrapfigure}
%In the 3GPP simulation models \cite{threeGPP_2015}, the NLOS bias, $\gamma_{ji}$, is typically assumed to be Gaussian distributed with a known mean and variance %given by the power delay profile and identically distributed for all AP. 
%and the APs are placed on an hexagonal grid. The combination of a perfect hexagonal grid with an identical Gaussian model for $\gamma_{ji}$ leads to unbiased localization estimates in 3GPP simulations, even in the presence of realistic channel models. If the localization of the device can be repeated with several independent ToA measurements, the error in the simulation studies drops with the squared root of the number of trials. 

In real cellular networks and unlike in typical 3GPP simulation models \cite{threeGPP_2015}, the NLOS biases are not Gaussian distributed, the distribution is not identical for all APs and in repeated trials the bias does not significantly change. Also, the cellular network are a far departure from the hexagonal idealization, especially in metropolitan areas. This combination in real cellular networks severely biases the localization accuracy, therefore necessitating that we model each source of error--the errors are larger than simulation predictions and not reducible by repeated trials. 

%Finally, the miscalibration of the APs is considered to be a zero mean Gaussian distribution. This prior is not unrealistic for general network deployment, but, at a particular location, these miscalibration errors are fixed biases that cannot be integrated out by multiple measurements. They need to be accounted for and canceled out. 

\section{Probabilistic ToA model}

In Fig. \ref{fig1}, we depict the graphical model for ToA localization. In this model, we have combined the quantization error, the thermal noise and the clock error in a single unbiased variable with known variance, $\sigma_{clk}^2$%, because they are all unbiased and unimodal 
\footnote{We can combine these errors because they are not the main limiting factor in localization performance and an individual treatment does not lead to reduced estimated error. Even if this were the main error driver, reducing the quantization error and thermal noise is technologically viable and the main component of this error would be the clock synchronization. The synchronization of the clocks can be considered Gaussian distributed \cite{Lombardi02}.}. Specifically, the model is described as follows:
\begin{align} 
\text{ToA}_{j}|d_j,\tau,\delta_{T_j},\gamma_{j},\sigma^2_{clk}&\sim\mathcal{N}(\tau+d_{j}+\gamma_{j}+\delta_{T_j},\sigma^2_{clk})\label{ToA_N}\\
\gamma_{j}|\boldsymbol{\pi}_j\sim\text{Cat}(\boldsymbol{\pi}_j)\qquad&\qquad \boldsymbol{\pi}_j|\boldsymbol{\beta}\sim \text{Dir}(\boldsymbol{\beta})\\
%\boldsymbol{\pi}_j|\boldsymbol{\beta}&\sim \text{Dir}(\boldsymbol{\beta})\\
\delta_{T_j}\sim\mathcal{N}(0,\sigma_{\delta_T}^2)\qquad&\qquad \delta_{\bar{\x}_j}\sim\mathcal{N}(0,\sigma_{\delta_\x}^2)
%\delta_{\bar{\x}_j}&\sim\mathcal{N}(0,\sigma_{\delta_\x}^2)
\end{align}
where  $d_j=||\x-(\bar{\x}_j+\delta_{\bar{\x}_j})||$ and the prior over $\x$ and $\tau$ is defined in Subsection~\ref{priorx}. The prior over the NLOS bias is discussed in the following subsection.

\subsection{Prior over the non-line-of-sight bias}\label{prior_NLOS}

The NLOS bias is due to the multiple paths the signal can take to travel from the transmitter to the receiver. If the paths are sufficiently apart in time, we refer to them as resolvable mulipath. %, because we can get a ToA estimate at the receiver for each path and keep the first path as the LOS component. 
If the paths are not separable, then they are considered unresolvable multipath and we can only estimate a ToA for all the paths clustered together in time. The estimated ToA is a weighted sum of all the individual ToAs for all the unresolvable paths. The paths are resolvable if the time difference between them is larger than the inverse of the bandwidth of the signal. 
%In UWB localization, the bandwidth is typically larger than 1GHz, hence the unresolvable multipath is less than 0.3 meters. The UWB system only requires knowledge of which paths are LOS and which are not \cite{Guvenc09}. Typically, unresolvable multipath would only add a few centimeters to the ToA estimation of the LOS component. The distance to the APs in UWB systems is measured in tens of meters and typically it is straightforward to estimate if the communication from the AP is LOS. Unfortunately, the need to space the APs by tens of meters apart precludes UWS system from being widely deployable in metropolitan areas. 
%
In cellular LTE, the bandwidth of positioning signals is typically 10MHz and the unresolvable multipath can add tens of meters to the estimate of the ToA, even when the LOS path is present in the detected signal. 
%Additionally, in cellular LTE, the base stations are even further away (several hundred of meters to kilometers), which might prevent the LOS path from being present at the received signal and the first arriving path would have added NLOS bias. 
%
%These differences explain why UWB localization systems can provide \textit{cm}-level localization performance, while cellular provides error that are orders of magnitude larger. Not only is the estimation error of the ToA is much coarser for lower bandwidths, but the unresolvable multipath severely affects the estimated ToA. While simple NLOS bias models for UWB localization system work well in practice \cite{Dardari09}, these models are not applicable to ToA localization in cellular environments and we need to characterize the NLOS bias to be able to provide accurate localization estimates.

To model the NLOS bias distribution, we use a discrete random variable. We assume that $\gamma_j$ can take the values $\ell\sigma_{clk}/10$ for $\ell=0, 1, \ldots$ with probability $\pi_{j\ell}$. The spacing of $\sigma_{clk}/10$ is chosen to be somewhat arbitrary. It just needs to be small enough with respect to the clock standard deviation so the discreteness of the distribution is not a limiting factor.
In our experiments, we have placed a fixed distribution over $\boldsymbol{\pi}_j$, which is based on the standard behavior of NLOS biases \cite{Medbo09}, that allows us to provide accurate localization directly.\footnote{Alternatively, we could have placed a Dirichlet prior over all $\boldsymbol{\pi}_j$, or even a continuous distribution (e.g. gamma) or mixture models (e.g. hierarchical Dirichlet processes \cite{Teh06}). However, for our experiments we lack of sufficient data to estimate the NLOS bias with these proposed priors. More data would allow us to learn the distribution over $\boldsymbol{\pi}_j$ and obtain more accurate estimates.} We have set:
\begin{equation}
\boldsymbol{\pi}_{j\ell}=\begin{cases} \frac{1}{2K},&  0\leq\ell<K\\
\frac{L-\ell+K-1}{L(L-1)},& K\leq\ell<L+K\\
0,& \ell\geq L+K
\end{cases}
\end{equation}
This model assumes that 50\% of NLOS biases are between 0 and $K$, which models the effect of the unresolvable NLOS bias over the LOS component. In the other 50\% of the other cases, the LOS path is not present and the NLOS bias can be as large as $L$.  In LTE, if we set $\sigma_{clk}$ to 10 meters, then $K=30$ assumes that the unresolvable NLOS bias can be as large as 30 meters, and setting $L=2000$ assumes the NLOS bias can be as large as 2 km, because the spacing in $\boldsymbol{\pi}_j$ is measured in intervals of $\sigma_{clk}/10$.

\subsection{Prior over the position of the object} \label{priorx}

%In our probabilistic model, we have not explicitly indicated a prior over $\mathbf{x}$ and $\tau$, but this prior is fundamental for accurate localization. Our prior is one of the key aspects of our proposal, together with the flexible hyper-prior over the NLOS bias. First, we propose a prior for standalone localization when only a single ToA measurement is available. We also use a Kalman filter \cite{Dardari15} over $\x$ and $\tau$ when there are repeated measurements in short periods of time that allows us to track the device (this model is available in the Supplementary Material). 
We need an effective prior distribution over $\mathbf{x}$ and $\tau$, because the NLOS bias can be as large as hundreds of meters. If the distribution over the NLOS bias allows for hundreds of meters and we are using TDoA localization in which $\tau$ is unknown, an unrestricted prior over $\mathbf{x}$ and $\tau$ would assign high probability to solutions far from where the APs are situated and, in many cases, outside the convex hull of the APs.

Any prior directly over $\mathbf{x}$ and $\tau$ may end up being too restrictive or too wide if it also depends on the APs that have already been heard. 
%This is because we might hear APs that are very far or we might force the solution to be too close to the serving AP (the one the object hears first). The prior would also need to be adjusted for each scenario to contemplate all the potential positioning of the object. 
This restricts the universal applicability of this procedure. Our adopted solution is simple and quite effective. Instead of putting a prior over  $\mathbf{x}$ and $\tau$, we place a prior over $d_{j}$. We assume, \textit{a priori}, that $d_{j}$ is uniformly distributed over the positive real numbers though other priors could be placed on this quantity as well \cite{coluccia2014rss}.

As $d_{j}$ measures the range (in polar coordinates) from the object to the APs, we penalize positions that are further away from the APs that have been heard by the object when we translate this distance into Cartesian coordinates. This means that the localization is typically within the convex hull of the APs, but it can also be outside of it if needed. With this prior, the posterior over $\mathbf{x}$ and $\tau$ becomes:
%\begin{align}\nonumber
%p(&\mathbf{x},\tau |\text{ToA}_{1},\ldots,\text{ToA}_{N},\bar{\mathbf{x}}_1,\ldots,\bar{\mathbf{x}}_N,\delta_{T_j}, \sigma_{clk}^2,\boldsymbol{\pi})\\\propto
%   &\prod_{j=1}^N \sum_{\ell=0}^{L} \pi_{j,\ell} \frac{\exp\left(-\frac{(\text{ToA}_{j}-\tau-d_j-\delta_{T_j}-\ell\sigma_{clk}/10)^2} {2\sigma_{clk}^2}\right)} {\sqrt{2\pi\sigma_{clk}^2}(d_j+\sigma_{clk})}\label{post}.
%   \end{align}
\begin{align}
%p(\mathbf{x},\tau |\text{ToA}_{1},\ldots,\text{ToA}_{N},\bar{\mathbf{x}}_1,\ldots,\bar{\mathbf{x}}_N,\delta_{T_j}, \sigma_{clk}^2,\boldsymbol{\pi})\propto
p(\mathbf{x},\tau |-)\propto
\prod_{j=1}^N \sum_{\ell=0}^{L} \pi_{j,\ell} \frac{\exp\left(-\frac{(\text{ToA}_{j}-\tau-d_j-\delta_{T_j}-\ell\sigma_{clk}/10)^2} {2\sigma_{clk}^2}\right)} {\sqrt{2\pi\sigma_{clk}^2}(d_j+\sigma_{clk})}\label{post}.
\end{align}

$d_j$ appears in the denominator when we transform the uniform prior for $d_j$ into the Cartesian coordinate system over $\mathbf{x}$. We add $\sigma_{clk}$ to avoid the singularity at the AP locations. In most cases, $d_j\gg \sigma_{clk}$ and this value is only to avoid numerical instabilities. To perform posterior inference in this model, we rely on an expectation propagation procedure detailed in the following section.
%To get samples from the posterior we either rely on a Metropolis-Hastings procedure \cite{MacKay03} with a Gaussian proposal of variance $\sigma_{clk}^2$ or an expectation propagation procedure detailed in the following section.

\section{Inference}\label{inference}
Expectation propagation (EP) \cite{Minka01a,Seeger05,Seeger08,Opper2005} is a technique in Bayesian machine learning for approximating posterior beliefs with a simpler set of exponential family distributions. 
% \cite{Minka01a,Seeger05,Seeger08,Opper2005}
%The ultimate goal is that the moments of the approximation match those computed from the original distribution, known as the moment-matching criterion. 
%Decompose the prior term  $p(\mathbf{x},\tau)$ in  \eqref{post}  as $p(\mathbf{x}|\tau)p(\tau)$. In the following, we develop an EP-based Gaussian approximation to 
%\begin{align}\label{joint2}
%p(\mathbf{x} |\tau,&\text{ToA}_{1},\ldots,\text{ToA}_{N},\bar{\mathbf{x}}_1,\ldots,\bar{\mathbf{x}}_N,  \sigma_{clk},\mathbf{p})\propto \prod_{j=1}^N \sum_{\ell=0}^{L}  p_{j\ell} \frac{1}{d_j}\frac{\exp\left(-\frac{(c(\text{ToA}_{j}-\tau)-(d_j+\mu_{jk}))^2}{2\sigma_{jk}^2}\right)}{\sigma_{jk}\sqrt{2\pi}}
%     % p(\text{ToA}_{j}|\tau,d_{j},\mu_{jk},\sigma_{jk}).\nonumber
%    \end{align}
%We assume that the prior term $p(\mathbf{x}|\tau)$ is non-informative, constant in the domain $\x\in\mathbb{R}^3$. We later discuss the case where this prior term follows a different distribution. 
Consider a family of Gaussian distributions which factorizes as follows:
\begin{align}\label{EPq}
q(\x)&=\frac{1}{\ZEP}\prod_{j=1}^{\N}\exp(\g_j^T\x-\frac{1}{2}\x^T\S_j\x),
\end{align}
where $\g_j\in\mathbb{R}^{\N}$ and $\S_j$ is a $3\times 3$ symmetric positive semi-definite matrix. Note that given $(\g_j,\S_j)$, $j=1,\ldots,N$, $q(\x)$ is a Gaussian distribution with covariance matrix and mean given by
\begin{align}\label{EPmoments}
\vSigma_{\EP}=\left(\sum_{j=1}^{\N}\S_j\right)^{-1}, ~\vmu_{\EP}=\vSigma_{\EP}\left(\sum_{j=1}^{\N}\g_j\right).
\end{align}

Our aim is to find  $\underline{\boldsymbol{\S}}^*=[\S^*_j]_{j=1}^{N}$ and $\underline{\boldsymbol{\gamma}}=[\g^*_j]_{j=1}^{N}$ such that $ \left(\vmu_{\EP},\vSigma_{\EP}\right)$ in \eqref{EPmoments} are an accurate estimate to the mean and covariance matrix of $\x$ w.r.t. the distribution in \eqref{post}. In the supplementary material, we describe an algorithm that iteratively refines the parameters vectors $\g_j$, $\S_j$ independently, so the update can be parallelized over $j=1,\ldots,N$. Our experiments have shown that the resulting algorithm  typically converges to a stationary point in a few iterations, where the complexity is $\mathcal{O}(N)$ per iteration.  Also, convergence is independent of the EP initialization, so we simply initialize $\g_j$ and $\S_j$, $j=1,\ldots,\N$ so that  $\vmu_{\EP}$ is at the center of the scenario and $\vSigma_{\EP}$ is a diagonal matrix with individual variance equal to a few meters.

\begin{figure*}[h]
	\centering
	\subfigure[]{\includegraphics[width=.24\textwidth]{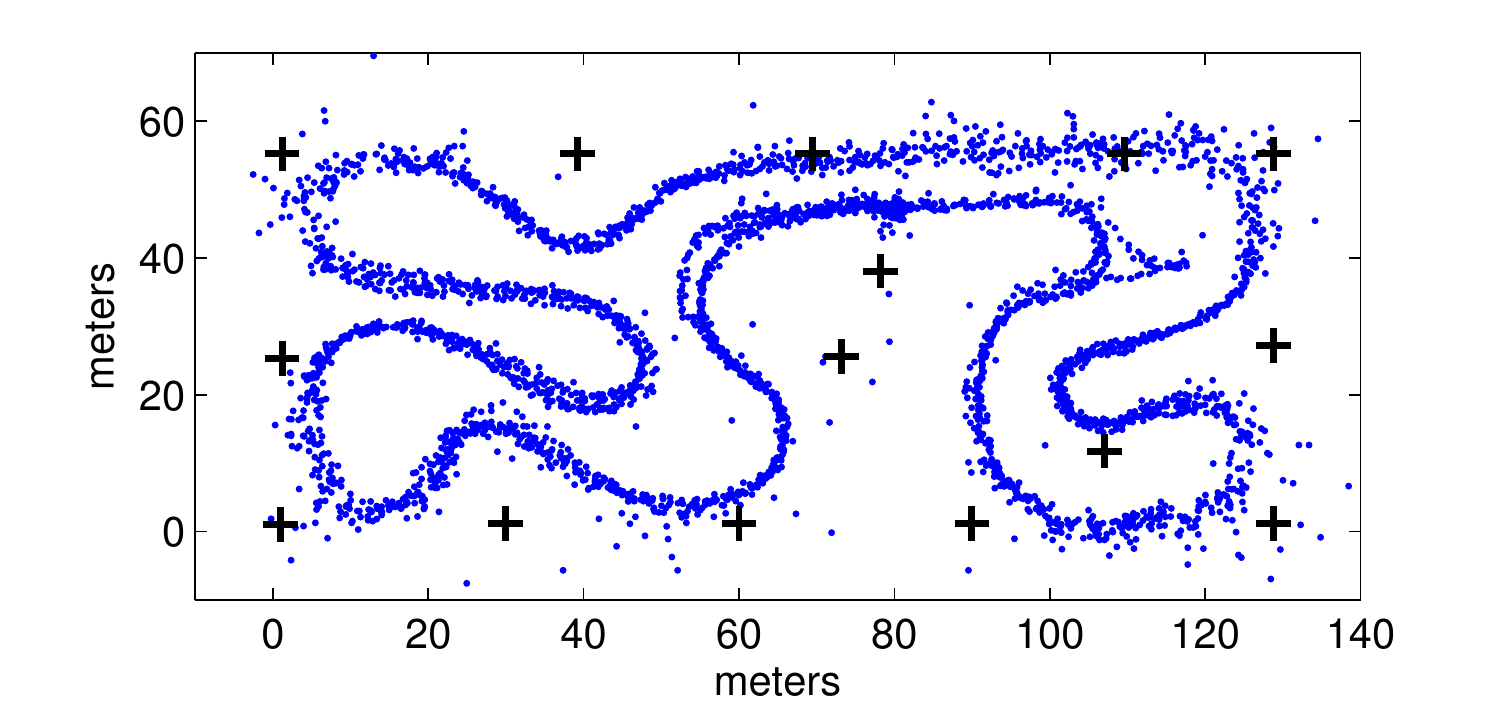}\label{fig11a}} 
	\subfigure[]{\includegraphics[width=.24\textwidth]{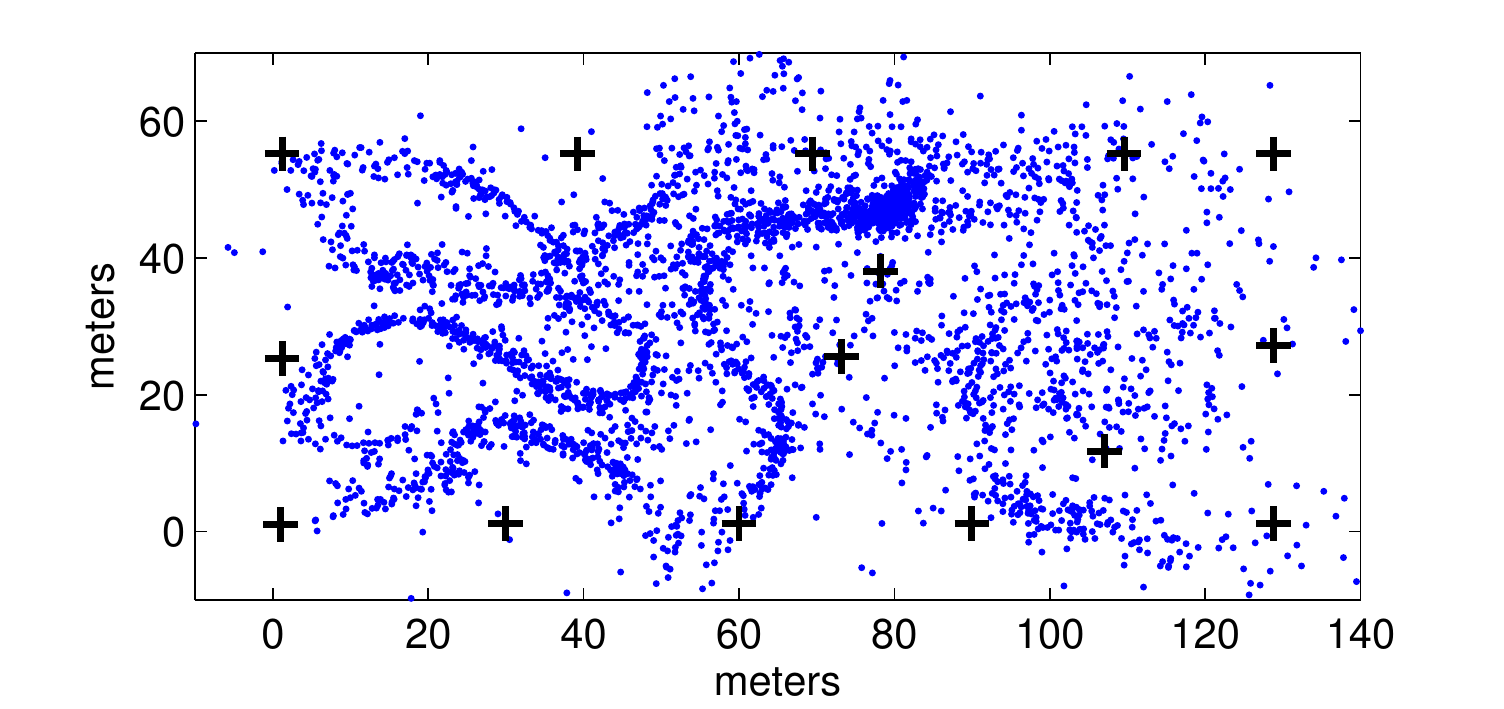}\label{fig11b}}
	\subfigure[]{\includegraphics[width=.24\textwidth]{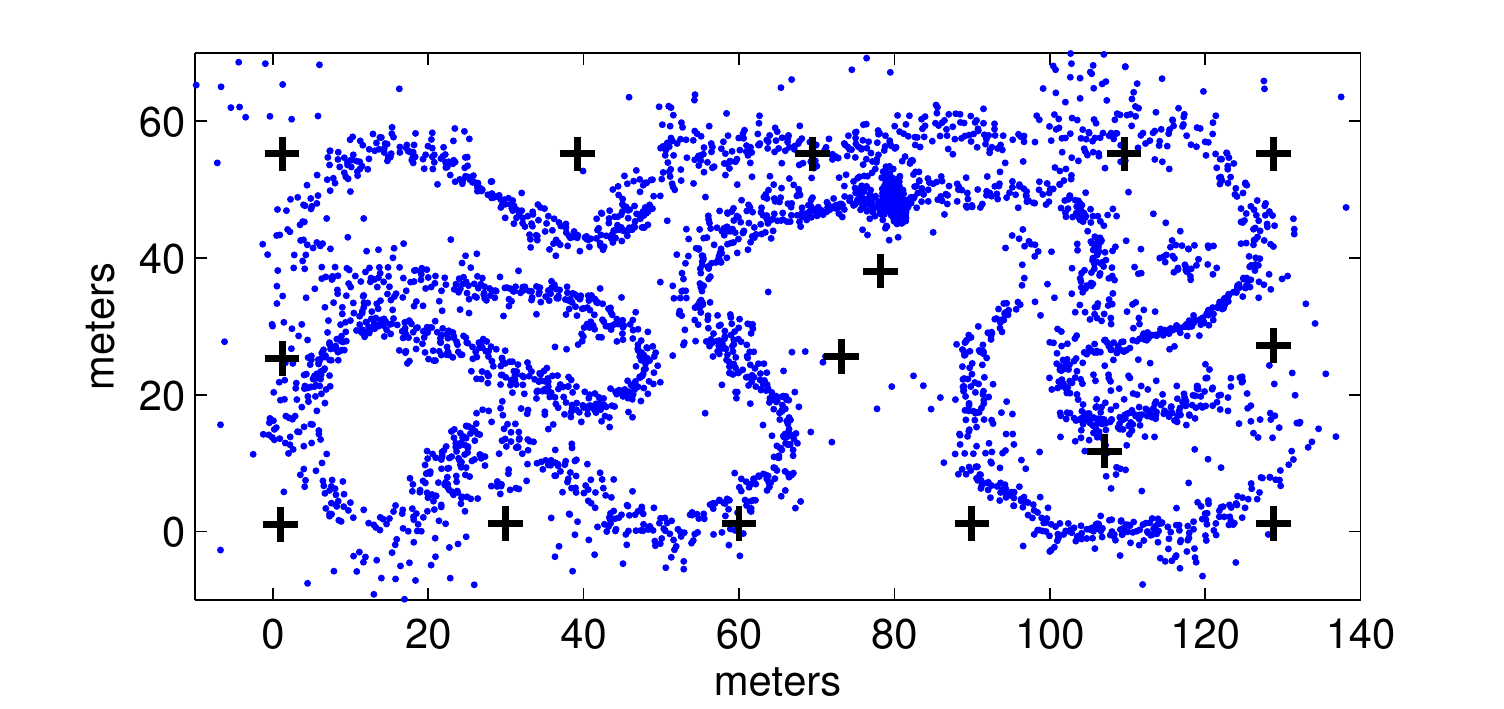}\label{fig11c}}
	\subfigure[]{\includegraphics[width=.21\textwidth]{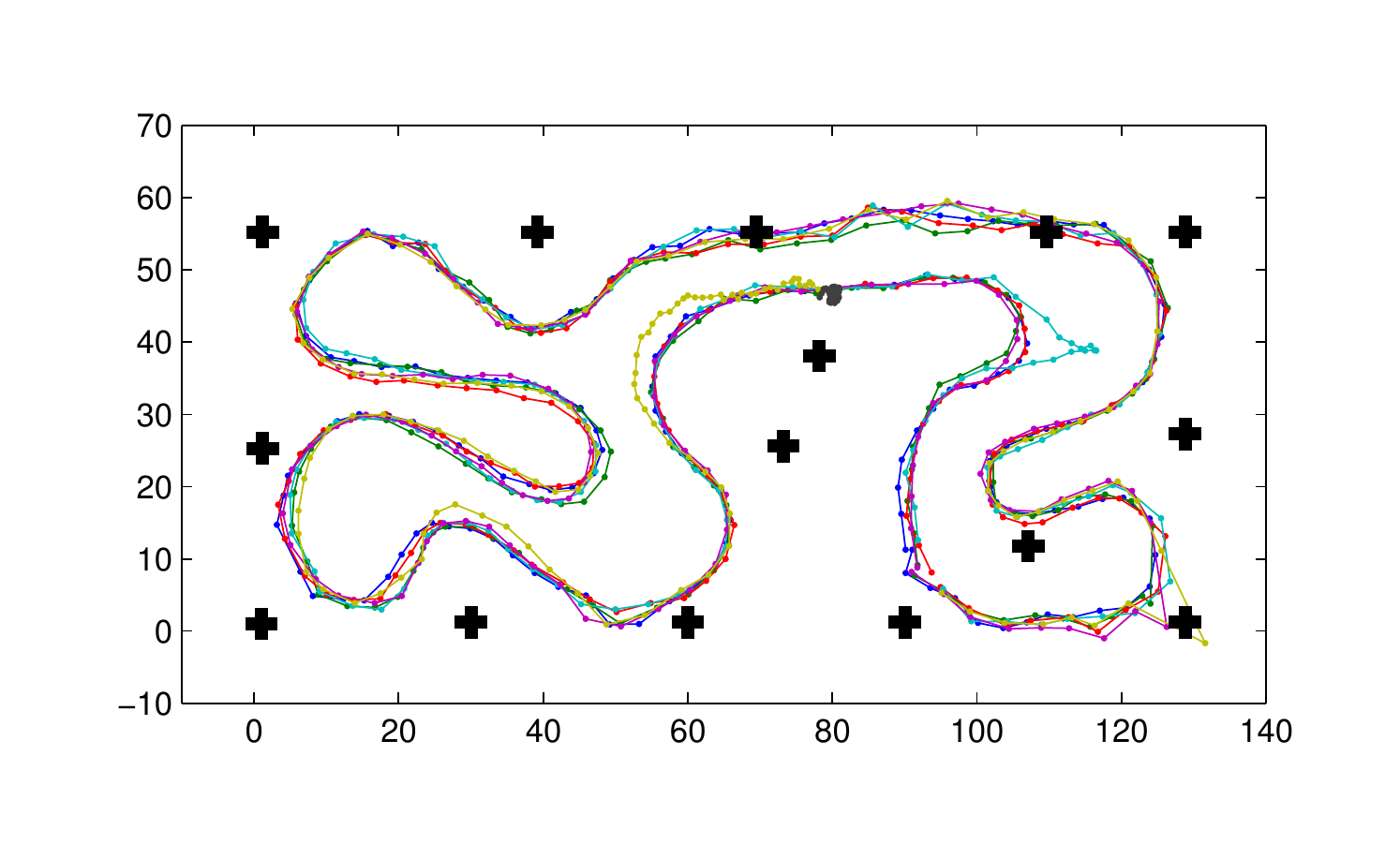}\label{fig11d}}
	\vspace*{-10pt}
	\caption{Raw localization for the 4650 transmissions for the second cart in (a) the performance of the proposed probabilistic algorithm. In (b) we show the least squares solution \cite{Guvenc09} and in (c) the nonlinear solution \cite{Fischer14}. In (d) estimated position of the second cart using our probabilistic algorithm with a Kalman filter.}
	\label{fig11}
\end{figure*}

\section{Experiments}\label{experiments}
%We have tested our algorithms in two different scenarios: a standalone localization network and a life LTE network in Atlanta and Chicago. The standalone network measures the ToA every 100ms and we have enough number of samples to test our Kalman filter and we have used this data to inspere a suitable model for $\boldsymbol{\pi}_j$, as introduced in Subsection \ref{prior_NLOS} and detailed in the supplementary material. In LTE we get a ToA measurement every 20-30 seconds, because PRSs are transmitted every 0.16 seconds and the Qualcomm chips waits for 100 transmissions before reporting a ToA estimate. The bandwidth of the PRS signal in LTE is 10MHz. For the LTE data, we report standalone localization estimate for each ToA measurements without tracking. We compared our proposed algorithm to the linear solution in \cite{Guvenc09} and the nonlinear solution is the method from \cite{Fischer14}.
We have tested our algorithms using over-the-air ToA measurements for two different scenarios. The first scenario is a proprietary wireless network installed in an indoor go-cart track. The second scenario is a live LTE cellular network. We compare our proposed algorithm to the linear solution from \cite{Guvenc09} and the nonlinear solution from \cite{Fischer14}.

\subsection{Indoor Go-Cart Track}

%We have received from Nanotron a data from an indoor go-kart circuit in which a kart goes round the circuit 6 times. We have about 4650 transmissions (7.5 minutes). We do not have the ground truth nor do we know the circuit layout, but we still are able to make a compelling case about the position on the karts. Nanotron has a proprietary synchronization algorithm that synchronizes the AP clocks with a standard deviation error below 1 meter. They use an 80MHz chirp signal in the 2.4GHz ISM band and the duration of their pulse is 4 $\mu$seconds. We have set $\sigma_{clk}=1$ meter, $K=40$ and $L=1500$, we discuss the setting of these values in the Supplementary Material.
The first scenario consists of ToA measurements taken in an indoor go-cart race track approximately of area 120m $\times$ 80m. The location-awareness startup company Nanotron, % (\texttt{www.nanotron.com}) 
in cooperation with go-cart system specialist SMS-Timing,  %(\texttt{www.sms-timing.com}) 
installed Nanotron's proprietary ToA localization system consisting of 15 APs. Two RF tags are attached to separate go-carts, and the two go-carts were driven simultaneously multiple times around the track. The system is an uplink system, where each go-cart tag transmits a known signal pulse of duration 4 $\mu$seconds and bandwidth 80MHz in the 2.4GHz unlicensed band, once every 100ms. The ToA is estimated at the APs. However, due to the limited range of the radios and interference, not all pulses are detected by all access points. The APs are synchronized through a proprietary Nanotron algorithm to achieve a standard deviation error below 1 meter. We have set $\sigma_{clk}=1$ meter, $K=40$ and $L=1500$.

The measurements from one go-cart are used to estimate the NLOS bias distributions, and the measurements from the second are used to measure the performance of the localization algorithms, as shown in Fig. \ref{fig11}(a). We note that we do not have the ground truth nor do we know the track layout, but we still are able to make a compelling case about the position on the karts based on the qualitative results of our model. We show the results for the linear algorithm \cite{Guvenc09} in (b), for the nonlinear algorithm \cite{Fischer14} in (c), and for our probabilistic ToA algorithm in (a). For the linear algorithm, the circuit is not visible in the right hand side of the plot and the nonlinear algorithms looks like a nosier version of the solution proposed by our algorithm, in which the circuit is outlined accurately. For our algorithm only the top left straight path looks a bit noisier. The larger errors occur when the device has only been heard by fewer than 6 access points out of the 15 that are available.

We show the result for our probabilistic algorithm with a Kalman filter in Fig. \ref{fig11d}.  In this plot, we have joined the dots with lines of different colors for each loop around the circuit. In this plot, we can see that our algorithm clearly traces the cart going around the track. It also reveals detailed information about the driver performance. The driver's lap time improves for the first three laps, but on the fourth lap (cyan) the cart veers off the track momentarily. 
%The fifth lap is the fastest, and the sixth lap (yellow) follows at the end a different path as the driver finishes. At the end there are 1250 samples shown in black in which the cart does not move, and we can see that the localization error is quite stable. 
%This detailed information is not visible with the other two algorithms even after processing them with a Kalman filter.

\subsection{Live LTE Networks}

%We have carried two experiments. A one test in Atlanta with improved hearability to show how our algorithm is able to improve once it hears more APs. A second experiment is a proof of concept in Arlington Heights (Chicago), in which we can show that estimating the biases of the APs can provide a much better localization performance. We have set $\sigma_{clk}=20$ meters, $K=10$ and $L=1000$.
We present performance results based on ToA measurements taken from live LTE networks. Two sets of data are studied, each corresponding to a different metropolitan area in the United States. The first set shows how the performance of the probabilistic algorithm improves when more APs are heard (i.e., as $N$ increases), and the second set shows how estimating the biases of the APs can further improve the accuracy of the algorithm. In both cases, we set $\sigma_{clk}=20$ meters, $K=10$ and $L=1000$. The results for the second LTE network are reported in the Supplementary Material. In this second example we show how $\delta_{T_j}$ can be learnt in order to significantly reduced the localization biases.

\subsubsection{LTE Network A}

For the first set of data, measurements are taken at 14 indoor locations within buildings around the downtown area of a large city. Two network configurations were employed: a standard configuration, and an enhanced configuration where the transmissions of the PRSs are coordinated to reduce interference and increase the number of APs heard. In the standard configuration, the average number of APs heard for a given transmission opportunity was $N=5$, and that value increases to 12 for the enhanced configuration.

\begin{figure}[h]
%\begin{wrapfigure}{R}{0.4\textwidth}
	\centering
	\includegraphics[width=.5\textwidth]{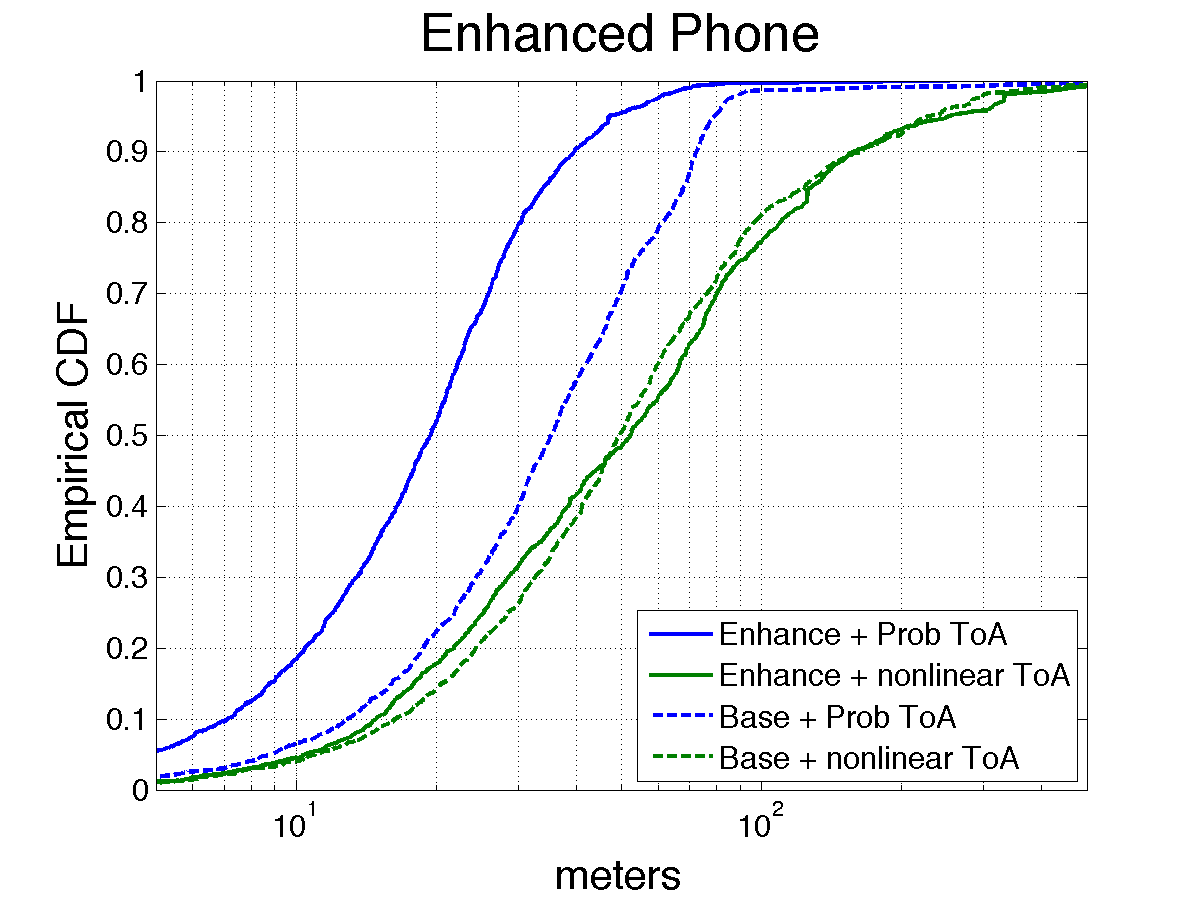}
%	\vspace*{-10pt}
	\caption{Empirical CDF localization error for an LTE cellular network  over 14 indoor locations using 2 different networks configurations (solid versus dashed), and two localization algorithms (blue and green).}
	\label{fig_At}
%\end{wrapfigure}
\end{figure}

The ground truth is known precisely, so it is possible to measure the error of the localization algorithms. In Fig. \ref{fig_At}, we show the empirical cumulative density function (CDF) for the localization error, in which the errors are sorted in the x-axis and the y-axis shows the proportion of transmissions that are below that error. For a given network configuration, 100 measurements are taken for each of the 14 locations, so each CDF curve consists of about 1400 points. For each network configuration, we consider the performance of the nonlinear algorithm and the proposed probabilistic algorithm. 

We observe that the enhanced network does not change the performance of the nonlinear algorithm significantly. At the 90th percentile, the baseline algorithm performance is about 150m. The performance of the proposed probabilistic algorithm under the standard network configuration is uniformly better than that of the nonlinear algorithm.  The 90th percentile error is reduced by about a factor of 2 to 72 meters. Increasing AP hearability with the enhanced network further reduces the error to 40 meters. Hence the probabilistic algorithm with the enhanced network operation improves the performance by a factor of 4 compared to the current baseline. The probabilistic algorithm achieves these gains because its NLOS model is able to detect the APs suffering from a strong NLOS bias and to reduce their effect in the location estimate.

\section{Conclusions}

%We have shown that time of arrival localization suffers from biases in their measurements and AP locations that prevent accurate localization with cellular technologies. We have proposed a probabilistic algorithm that it can learn the biases for the APs for fixed locations and learn the NLOS bias to significantly improve localization to below meter level localization. These improvements can only be realized once the geo-localization problem is cast as a probabilistic machine-learning problem in which all sources of error are accounted for and for which we can learn from measurements. All our results are based on real data from life localization and LTE networks. With the enhancement coming to 5G communication standards, with more hearibility and more resolution, we should expect that AP compensation and NLOS learning will bring ToA localization as the {\em de facto} tool for cell phones and IoT devices. 

We have proposed a probabilistic algorithm that can learn the biases corresponding to the two most significant sources of error in time-of-arrival localization using cellular networks. We then develop a novel multilateration algorithm by casting the localization problem as a probabilistic machine-learning problem in which all sources of error are accounted for. Using measurements from a live LTE network, we have shown that the proposed algorithm reduces the error by an order of magnitude--to less than 10 meters. 
%Enhancements in future 5G cellular standards could result in improved measurement quality and additional performance gains. Therefore with indoor and outdoor coverage by cellular signals and GPS-level accuracy, 5G cellular localization could become the \textit{de facto} technology for enabling scalable, universal, and accurate tracking of future handset and IoT devices.

%\bibliography{example_paper}

\newpage

\bibliographystyle{IEEEtran}
\bibliography{EP,Localization}
\newpage

\section*{Supplementary Material}

\subsection{EP Iterative refinement}
We  first describe the update of the pair $(\g_m,\S_m)$ assuming that  $\g_j$, $\S_j$, for  $j\neq m$ are fixed. The main idea is to remove the $m$-th factor, $\exp(\g_m^T\x-\frac{1}{2}\x^T\S_m\x)$, from $q(\x)$ in \eqref{EPq} and include the associated ``true'' factor. The pair $(\g_m,\S_m)$ is updated according to the moments computed in this modified distribution. We define the \emph{cavity distribution} as
\begin{align}
\cav=\frac{1}{\hat{Z}^{\backslash m}}\prod_{\substack{j=1\\j\neq m}}^{\N}\exp(\g_j^T\x-\frac{1}{2}\x^T\S_j\x),
\end{align}
and \emph{the tilted} distribution as
\begin{align}\label{tilted}
\til &\propto \sum_{\ell=0}^{L} \frac{ \pi_{m,\ell}}{(d_m+\sigma_{clk})}\tili
\end{align}
where
\begin{align*}
\begin{split}
\tili&\propto\cav \exp\left(\frac{-(\text{ToA}_{j}-\tau-d_j-\delta_{T}-\ell\sigma_{clk}/10)^2} {2\sigma_{clk}^2}\right)
\end{split}
\end{align*}
%\begin{align}
%\begin{split}
%\tili&\propto\cav \times \\ 
%&\exp\left(-\frac{(\text{ToA}_{j}-\tau-d_j-\delta_{T}-\ell\sigma_{clk}/10)^2} {2\sigma_{clk}^2}\right)
%\end{split}
%\end{align}

%\\\nonumber%\frac{\cav}{\tilZ}  \left(\sum_{k=1}^{R_j} p(\text{ToA}_{m}|\tau,d_{m},\mu_{mk},\sigma_{mk})\right)
%&\propto\sum_{\ell=0}^{L} \frac{ \pi_{m,\ell}}{(d_m+\sigma_{clk})}  \cav \times \\
%&\exp\left(-\frac{(\text{ToA}_{m}-\tau-d_m-\delta_{T}-\ell\sigma_{clk}/10)^2} {2\sigma_{clk}^2}\right)\\
%&\propto\sum_{\ell=0}^{L} \frac{ \pi_{m,\ell}}{(d_m+\sigma_{clk})}\tili
%=\sum_{k=1}^{R_j} \pi_{mk} ~ \tili,
%\end{split}
%\end{align}
To update the pair $(\g_m,\S_m)$, we approximate the mean tilted distribution as follows: 
\begin{align}\label{exp1}
\E_{\til}\left[\x\right]&\approx\frac{\displaystyle \sum_{\ell=0}^{L}  \frac{ \pi_{\ell}}{(d^{\ell}_m+\sigma_{clk})}  ~  \E_{\tili}\left[\x\right]}{\displaystyle \sum_{q=0}^{L} \pi_{q}/(d^{q}_m+\sigma_{clk})},%\label{exp2}
%\E_{\til}\left[\x\x^T\right]&\approx\frac{\displaystyle \sum_{\ell=0}^{L}  \frac{ \pi_{\ell}}{(d^{\ell}_m+\sigma_{clk})}  ~  \E_{\tili}\left[\x\x^T\right]}{\displaystyle \sum_{q=0}^{L} \pi_{q}/(d^{q}_m+\sigma_{clk})}
%\E_{\til}\left[\x\x^T\right]&=\sum_{i=1}^{\R}\frac{ p_{j\ell}}{d^{jk}_j} ~  \E_{\tili}\left[\x\x^T\right],
\end{align}
 $d^{\ell}_m$ is the Euclidean distance between the $m$-th AP and  $\E_{\tili}\left[\x\right]$ is the mean of $\tili$. Note that  we are replacing the $1/(d_m(\x)+\sigma_{clk})$ penalty term in \eqref{tilted} by its value at the mean of the inner distribution  $\tili$. We can proceed similarly to estimate the correlation of the tilted distribution $\E_{\til}\left[\x\x^T\right]$. Finally, the first two moments of $\tili$ are 
estimated by a Gaussian-quadrature approximation, in which we use a first order Taylor expansion of the term $d_{m}=\sqrt{||\mathbf{x}-\bar{\mathbf{x}}_m||^2}$ around the mean of the cavity distribution $\cav$, obtaining a Gaussian approximation to $\tili$.  
The updated pair $(\g^*_m,\S^*_m)$ is updated as follows:
%is chosen so that the moments of the distribution, $\cav \exp((\g_m^*)^T\x-\frac{1}{2}\x^T\S^*_j\x)$ (upon normalization), match the moments in \eqref{exp1}. A simple computation shows that:
\begin{align}
\S_m^*&=\left(\text{CoVar}_{\til}[\x]\right)^{-1}-\sum_{\substack{j=1\\j\neq m}}^{\N}\S_j\qquad\\
\g_{m}^*&:=\left(\text{CoVar}_{\til}[\x]\right)^{-1}\E_{\til}\left[\x\right]-\sum_{\substack{j=1\\j\neq m}}^{\N}\g_j.
\end{align}

\subsection{LTE Network B}

%In this section we describe an experiment in which we demonstrate that significant AP calibration errors indeed exist but they can be corrected to further improve the performance of the probabilistic algorithm. 
The experiment consists of two phases: a training phase where the calibration errors for the APs are estimated, and a testing phase where the probabilistic algorithm is run on the corrected ToAs.  The network operated with the standard hearability. %The map with the location of the training and testing sites is shown in the Supplementary Material.%t we need further experimentation to solve this problem correctly. In this experiment we use the standard network configuration and we can only use 7 APs to localize the devices. 
In the training phase, we took 100 measurements in a parking lot  in order to estimate the calibration bias of nearby APs. We made the following three assumptions: all the ToA measurements were LOS; the reference AP did not have a bias; and all the calibration bias was due to cable length (i.e. $\delta_{\x_j}=0$). 
%
%The first assumption is justified because there are no tall buildings in the area and we are just focusing on the close by APs (red dots in Figure \ref{position2}). The second assumption is justified because we observed that by adding a fixed temporal bias to all the APs, the estimated device location does not change. Because that offset will be absorbed in the estimate of $\tau$ (time of flight). The last assumption is justified, because  the training and test positions are nearby, and all of them see the APs from almost the same angle. This means that the bias that we are compensating might be a combination of $\delta_{\x_j}$ and $\delta_{T_j}$, and we will only be able to know which one it is with further testing. Nevertheless, at least one of errors is significant enough. 
In this training location, we got a clear GPS reading and its 
location as the ground truth to estimate $\delta_{T_j}$ for each of the visible APs. We found that most APs had an error between 0 and 50 meters. In Fig. \ref{position2}, we show the location of the APs. The ones in red numbered (1, 3, 5, 6, 7, 10 and 13) are the one that we see all the time in the training point. The other we see them infrequently and we cannot compensate their biases.

%\begin{figure*}[h]
%	\centering
%	\subfigure[]{\includegraphics[width=0.3\textwidth]{Testsites_AH.png}\label{}} 
%	\subfigure[]{\includegraphics[width=0.26\textwidth]{APs_location_AH.pdf}\label{}}
%	\vspace*{-10pt}
%	\caption{}
%	\label{}
%\end{figure*}

\begin{figure}[h]
\includegraphics[width=0.4\textwidth]{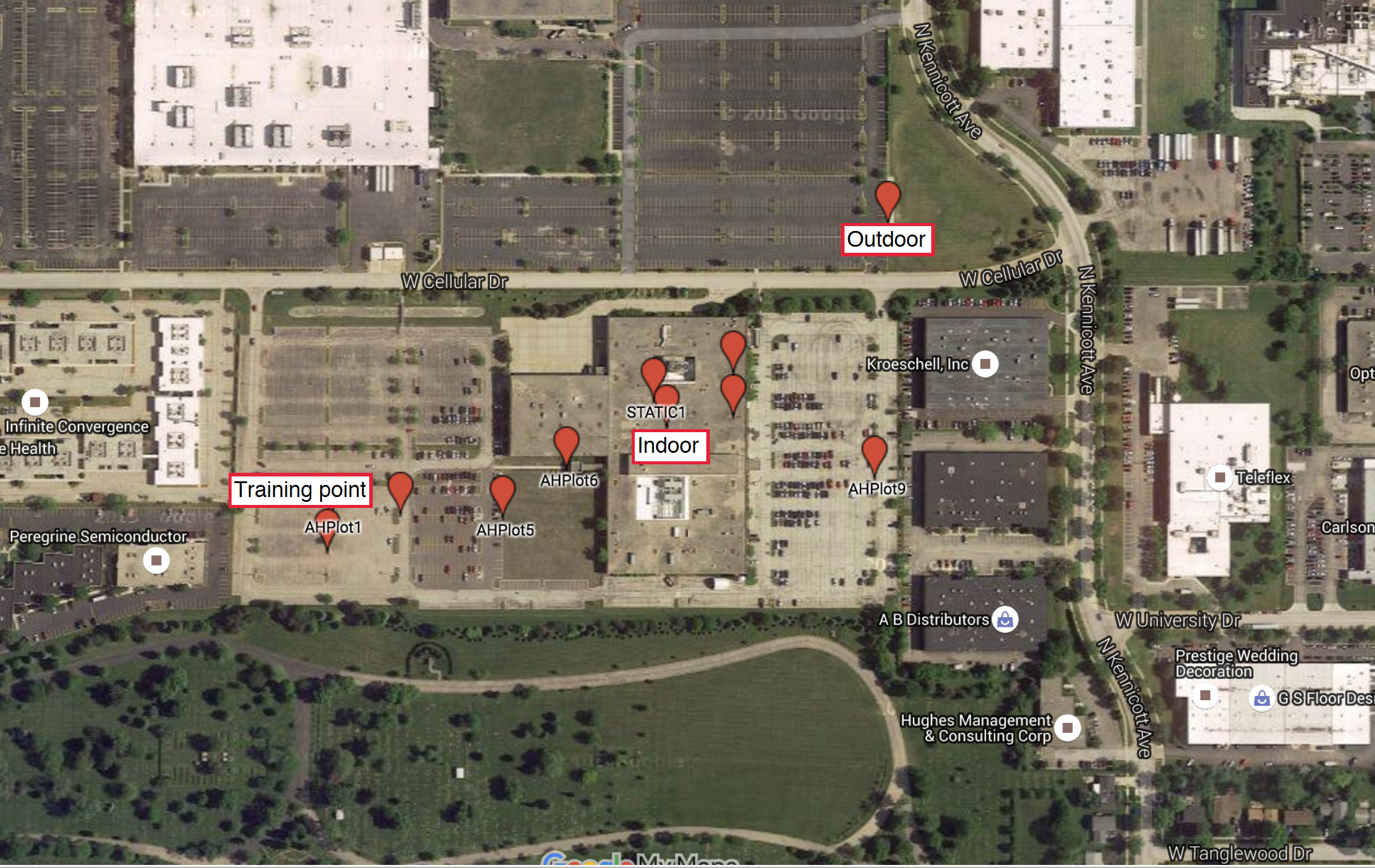}
\centering
\caption{Satellite map for the position of the training and test point for LTE Network B.}
\label{position1}
%\vspace*{-10pt}
\end{figure}

 % and we cannot use them for improved localization. %Once we increase the hearbility, we might be able to use them for improve localization in which we can compensate their position and cable length biases, besides having more APs to compensate the NLOS biases. 
%The green dots indicates the position of the training and test points.%, in which we can see that the angle to all the APs in very close, as argued previously. 
%The estimated biases for each of the APs are: $-3\times 9.77$ meters for AP 1, $1\times 9.77$ meters for AP 3, $-1\times 9.77$ meters for AP 5, $-5\times 9.77$ meters for AP 7, $-5\times 9.77$ for AP 10 and $1\times 9.77$ for AP 13. AP 6 is the reference AP and its bias is 0 meters. We measure the biases in order of 9.77 meters, because that it is the resolution of the LTE network.

\begin{figure}[h]
\includegraphics[width=0.25\textwidth]{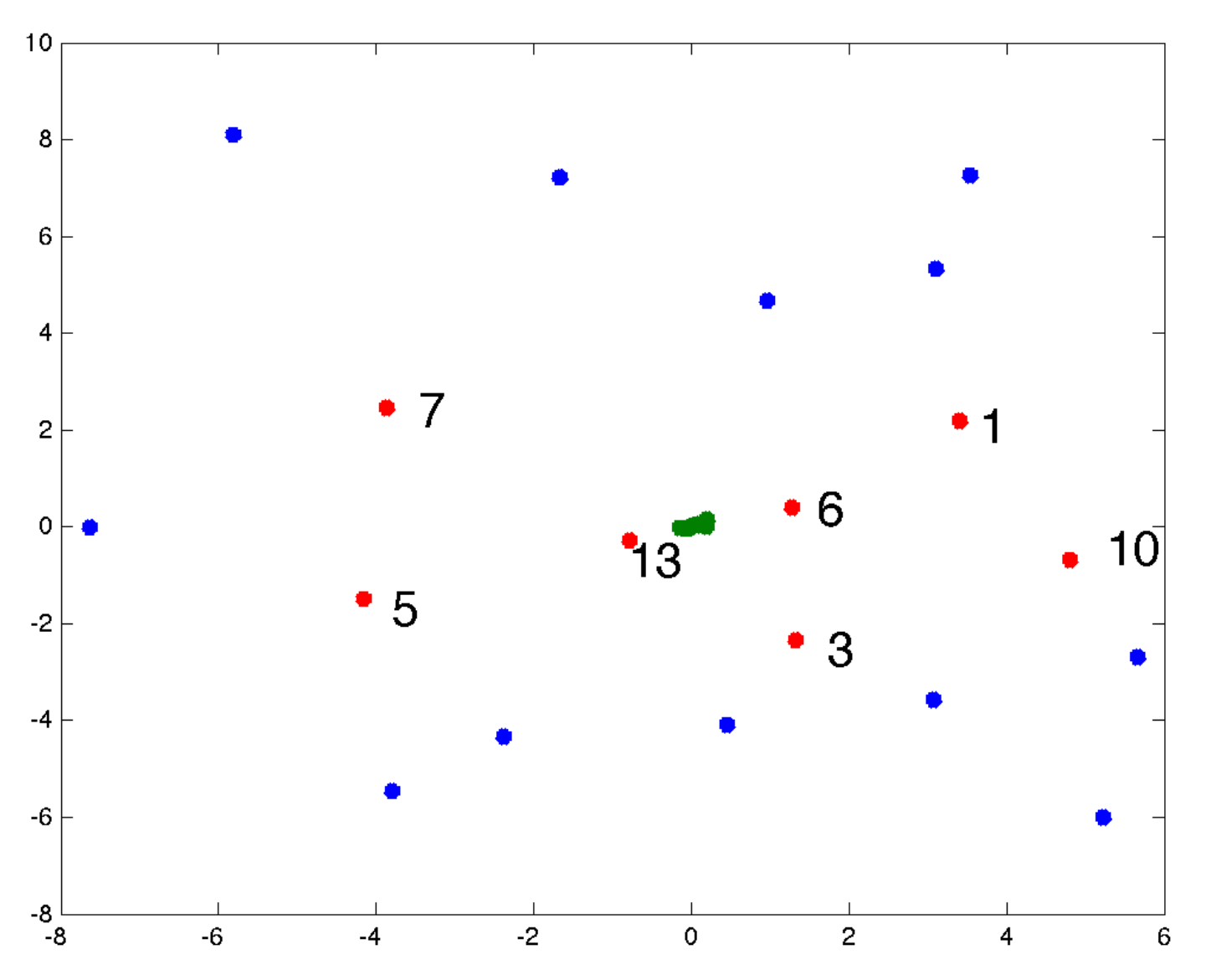}
\centering
\caption{Map for the position of the APs for LTE Network B. The green dots are the positions of the device from Fig. \ref{position1}.}
\label{position2}
%\vspace*{-10pt}
\end{figure}

%For the second LTE experiment, we show in Figure \ref{position1} the Satellite images of the physical locations. In which we can see the training point in the west parking lot and the outdoor testing point in the north-east parking lot. The indoor test point in on the ground floor in the middle of the building. It is not close to windows.

 \begin{figure}[h]
\centering
\subfigure[Outdoor]{\includegraphics[scale=0.35]{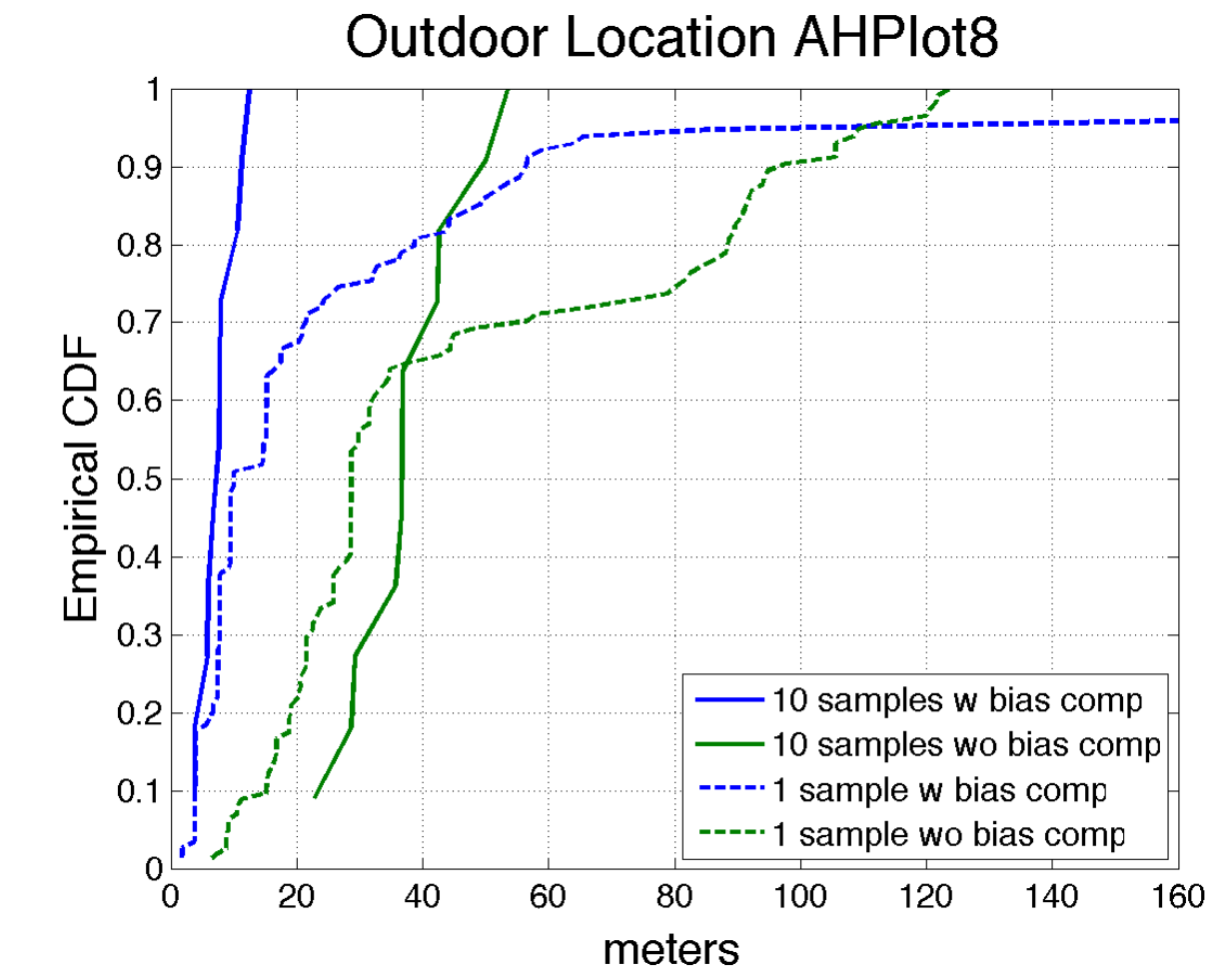}\label{LTE1a}}
\subfigure[Indoor]{\includegraphics[scale=0.35]{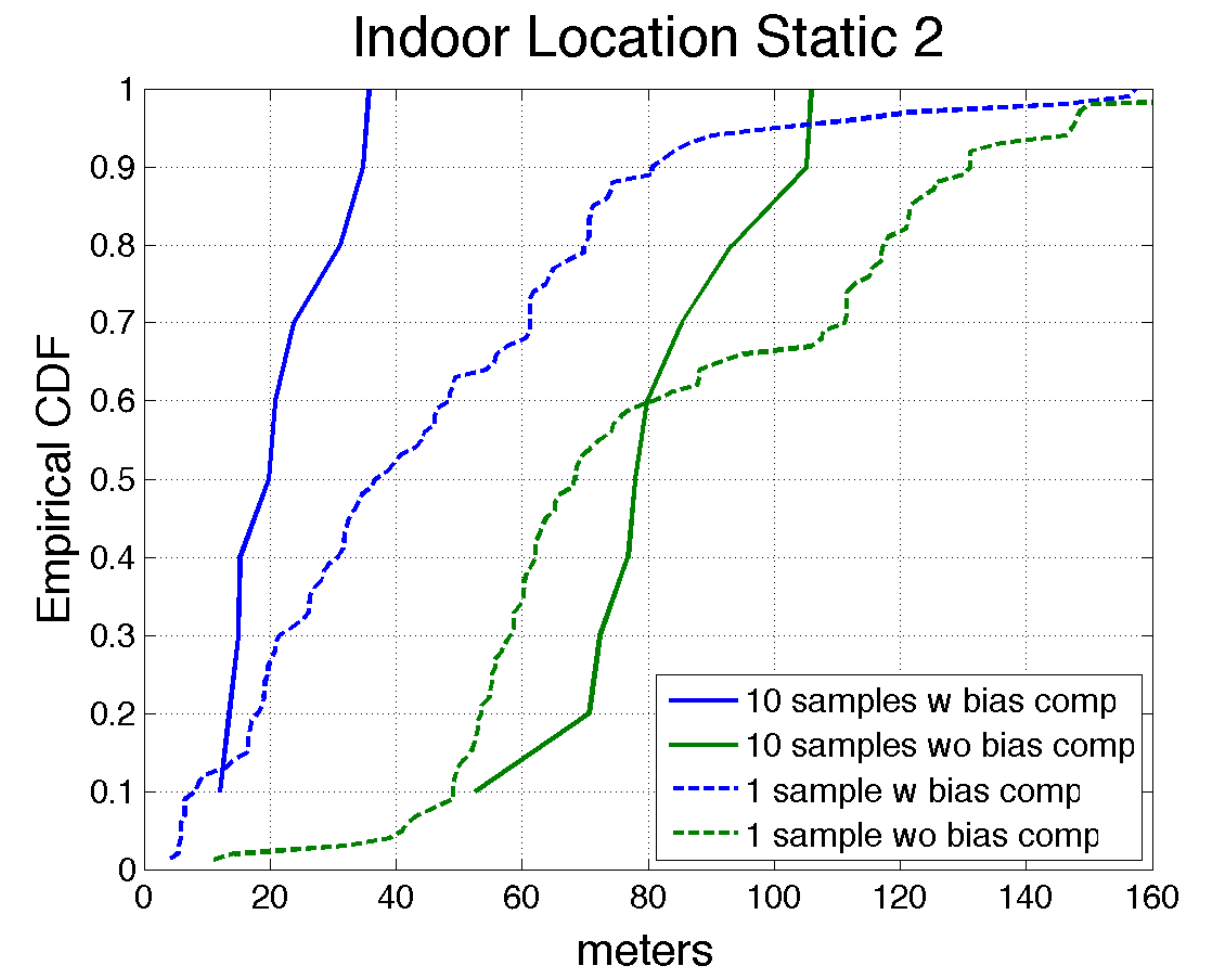}\label{LTE1b}}
\vspace*{-10pt}
\caption{Empirical CDF localization error for a 100 measurements outdoor (a) and indoor (b).}
\label{AH1}
\end{figure}

In the testing phase, we took additional measurements at two nearby locations. One was in another parking lot (Fig. \ref{position1} labeled as {\em Outdoor}), and the second one was inside a building (Fig. \ref{position1} labeled as {\em Indoor}). The ToAs for AP $j$ was corrected using the estimated calibration error $\delta_{T_j}$. The empirical CDF localization error for both locations is shown in Fig. \ref{AH1}. It can be seen that once we compensate for the biases, the localization errors reduce considerably. 
%The outdoor 90th percentile error is reduced from 50 meters to 10, and the indoor 90th percentile error is reduced from 105m to 35m. In the indoor location, we only hear 4 APs, and we do not have the necessary redundancy to further improve the localization performance. We expect that by turning on the network enhancement would dramatically improve the performance. We note that the outdoor error of 10m is comparable to GPS accuracy

%To validate these results we need to take further measurements, so we can distinguished between cable miscalibration and AP localization error. These additional measurements would also help discard if the effect that we are finding is a NLOS bias, because we will be seeing the APs from different angles. In any case, we do not believe that we are compensating NLOS biases, because the outdoor points are not close to buildings and are separated a few hundred meters and we have an indoor location that should suffer NLOS bias.

\end{document}